\documentclass[letterpaper]{article} 
\usepackage{iclr2025_conference,times}

\usepackage{amsmath,amsfonts,bm}

\def\eqref#1{equation~\ref{#1}}

\def\1{\bm{1}}

\DeclareMathAlphabet{\mathsfit}{\encodingdefault}{\sfdefault}{m}{sl}
\SetMathAlphabet{\mathsfit}{bold}{\encodingdefault}{\sfdefault}{bx}{n}

\usepackage{hyperref}
\usepackage{url}

\usepackage{verbatim}
\usepackage{algorithm}
\usepackage{algorithmic}
\usepackage{comment}
\usepackage{cite}

\usepackage{graphicx}
\usepackage{makecell}
\usepackage{booktabs}
\usepackage{subcaption}

\title{CritiPrefill: A Segment-wise Criticality-based Approach for Prefilling Acceleration in LLMs
}

\author{
    Junlin Lv, Yuan Feng, Xike Xie, Xin Jia, Qirong Peng, and Guiming Xie \\
AI Center, Guangdong OPPO Mobile Telecommunications Corp.,Ltd. \\
Guangdong, China\\
\texttt{\{junlin.lv, yuanfeng.yf\}@outlook.com, xkxie@hotmail.com}\\
\texttt{\{jiaxin, pengjirong, xieguiming\}@oppo.com}
}

\iclrfinalcopy
\begin{document}

\maketitle

\begin{abstract}
Large language models have achieved notable success across various domains, yet efficient inference is still limited by the quadratic computation complexity of the attention mechanism. The inference consists of prefilling and decoding phases.
Although several attempts have been made to accelerate decoding, the inefficiency of the prefilling phase, especially for long-context tasks, remains a challenge. In this paper, we observe a locality in query criticality during the prefilling phase of long-context processing: adjacent query tokens tend to focus on similar subsets of the past Key-Value (KV) cache. 
Based on this observation, we propose CritiPrefill, a criticality-based segment-wise prefilling method.
This method partitions the input sequence's queries and KV cache into segments and blocks, utilizing a segment-wise algorithm to estimate the query criticality. By pruning non-critical computations between query segments and cache blocks in the self-attention mechanism, the prefilling process can be significantly accelerated.
Extensive evaluations on multiple long-context datasets show up to 2.7x speedup on Llama3-8B and 3.0x speedup on Yi-9B for 128K context length on a single A100 GPU, with minimal quality degradation.
\end{abstract}

\section{Introduction}

Large Language Models (LLMs) have been widely applied across various fields, showcasing impressive capabilities in handling long-context tasks such as long text comprehension~\citep{laban2023summedits}, multi-turn dialogue QA~\citep{multi_hop1}, in-context learning~\citep{dong2022survey}, and agent tasks\citep{Wang_2024}. 
The rapid advancement of open-source LLMs has significantly reduced the training costs for downstream applications. However, a key challenge remains in the quadratic inference cost of the self-attention mechanism in Transformer layers while processing long sequences \citep{wan2024efficientlargelanguagemodels,dong2023survey,zhou2024survey}. 

Typically, LLMs are composed of multiple Transformer layers, each containing a self-attention layer that significantly contributes to the inference cost in long-sequence scenarios~\citep{wang2020linformerselfattentionlinearcomplexity}. Inference within each self-attention layer consists of two phases: the {\it prefilling phase} and the {\it decoding phase}.
During the prefilling phase, LLMs calculate the Key-Value (KV) cache for all input tokens and predict the first output token, which takes the majority of the computation cost during inference~\citep{mooncake}. In the subsequent decoding phase, the model generates tokens autoregressively by leveraging the most recent query token along with the entire KV cache from all previous steps.
While this phase involves less computation, it is primarily constrained by the input/output (I/O) latency of the KV cache~\citep{leviathan2023fast}, resulting in a memory-bound bottleneck.

\begin{figure}
    \centering
    \begin{minipage}{0.48\linewidth}
        \centering
        \includegraphics[width=\linewidth]{./Figures/time_ratio.pdf}
        \caption{Acceleration of CritiPrefill}
        \label{fig:time_ratio}
    \end{minipage}\hfill
    \begin{minipage}{0.48\linewidth}
        \centering
        \includegraphics[width=\linewidth]{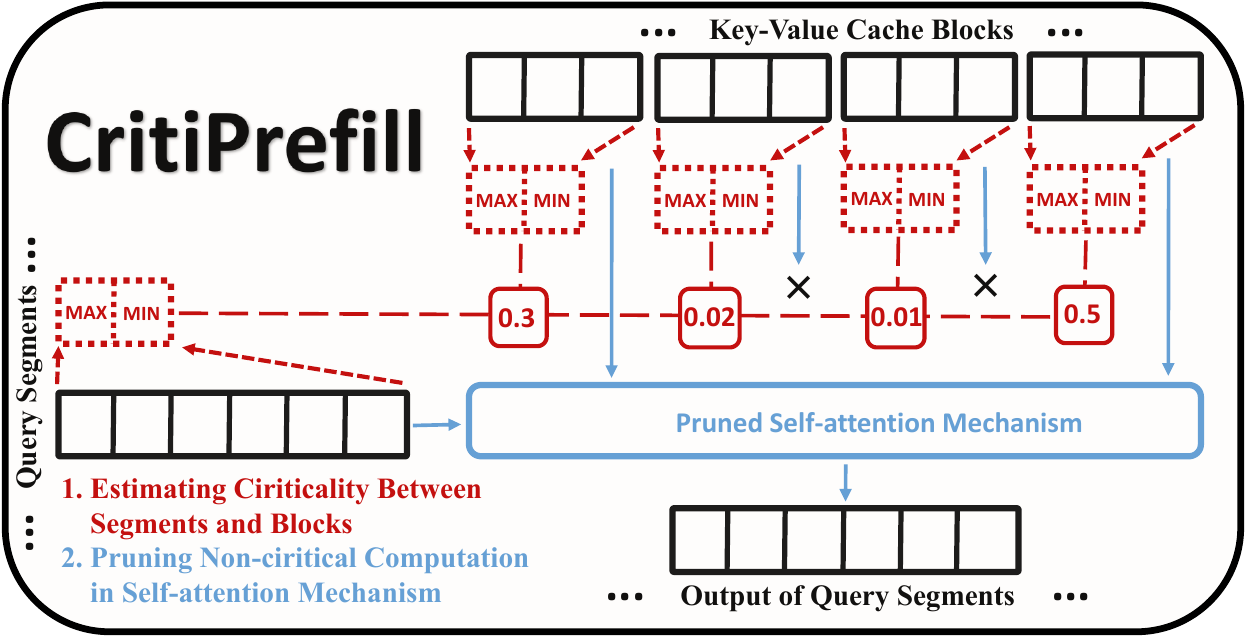}
        \caption{Framework of CritiPrefill}
        \label{fig:critiprefill}
    \end{minipage}
\end{figure}

To address this challenge, plug-and-play (P\&P) methods have been extensively explored for seamless integration into any LLM, to reduce inference costs without the need of substantial fine-tuning expenses \citep{fa1,pageattn,kachris2024surveyhardwareacceleratorslarge}.
Among those works, recent advances in efficient decoding~\citep{ada,SnapKV,h2o} have made remarkable progress,
especially by identifying key subsets of KV caches based on the last query token at each decoding step~\citep{quest}.
By pruning non-critical KV caches, these methods minimize redundant computations, significantly reducing cache I/O latency and accelerating the decoding phase. 

However, in typical long-sequence scenarios, such as long-context QA, the input sequence is often much longer than the output sequence, resulting in a significantly higher prefilling time than the total decoding time. 
As shown in Figure \ref{fig:time_ratio}, when the sequence length reaches 128K, the prefilling time can account for over 95.6\% of the total inference time, indicating that the main bottleneck in long-sequence tasks lies in the prefilling phase.
However, accelerating the prefilling phase remains a significant challenge due to its highly parallel yet computationally intensive nature.
Existing methods for prefilling acceleration typically require architectural changes~\citep{shen2021efficient,qin2024lightning,gla} or extensive fine-tuning~\citep{longformer,zaheer2020big,longllama}, which limiting their integration with pretrained LLMs \citep{deepspeed,megatron}.
This raises a critical question: {\it Can the P\&P strategy for critical KV cache identification, which has promise in the decoding phase, be extended to the prefilling phase?}

Nevertheless, it is non-trivial to identify critical KV cache.
The token-wise criticality estimation \citep{quest}, while effective in decoding, introduces significant computational overhead, making it impractical for the compute-intensive prefilling phase. 
However, we observe a key locality pattern: {\it although the critical subset of the KV cache varies with different query tokens, neighboring query tokens tend to share similar critical KV caches.}
Based on this observation, we propose the {\it CritiPrefill}, which leverages segment-wise query criticality estimation for pruning attention computation, thus accelerating the prefilling phase.

As outlined in Figure~\ref{fig:critiprefill}, the CritiPrefill partitions both query tokens and KV caches into fixed-size query segments and cache blocks, before the prefilling phase of each layer. 
A segment-wise algorithm then estimates the criticality of each cache block for each query segment. 
Additionally, inspired by inter-layer similarity in LLMs \citep{dai2019transformer, liu2023deja, zhang2023efficient}, we propose a layer-fusion mechanism to further refine the final criticality score.
By pruning non-critical computation between query segments and cache blocks in the self-attention mechanism, the CritiPrefill significantly accelerates the prefilling process. 
We evaluate CritiPrefill across various long-context QA tasks with different input lengths, showing substantial speedups with minimal accuracy loss. 
Additionally, tests on the ``Needle-in-a-Haystack'' task further confirm that CritiPrefill maintains information retrieval capacity while reducing prefilling time for long sequences, achieving up to a 2.7x speedup on Llama3-8B and 3.0x on Yi-9B using a single A100 GPU.

\section{Method}
\subsection{Query Criticality in Attention Mechanism}
Modern LLMs are generally constructed using multi-layer Transformer blocks, with the self-attention mechanism as the central operation. During the prefilling phase, the computation in each layer’s self-attention is heavily dependent on the hidden states $X \in \mathbb{R}^{n \times d}$ of all input tokens from the previous layer, which are transformed into Query $Q \in \mathbb{R}^{n \times d}$ , Key $K \in \mathbb{R}^{n \times d}$, and Value $ V \in \mathbb{R}^{n \times d}$ matrices as follows:
\begin{equation}
Q = W_Q X; \:  K = W_KX; \: V = W_VX
\end{equation}
The output $O$  is then computed using attention weights $A$:
\footnote{For clarity, we illustrate only a single head in the multi-head attention mechanism, which can be easily extended to multiple heads in practice.}
\begin{equation}
O = AV \: \text{where} \:A = \text{softmax}(\frac{QK^T}{\sqrt{d}})
\end{equation}

The matrix multiplications in self-attention mechanisms result in a quadratic computational cost relative to sequence length \( n \), which contributes to the significant computational load during the prefilling phase. In the decoding phase, LLMs use only the query of the last token \( q = Q[-1] \) along with cached KV pairs to autoregressively generate the next token.
Although the computational cost per decoding step is low, large sequence length \( n \) can lead to considerable KV cache I/O latency, creating a memory-bound bottleneck.
Fortunately, due to the inherent sparsity of attention weights~\citep{ge2023model}, only a small subset of critical KV cache significantly influences the output generation.
Thus, excluding non-critical portions of the KV cache from self-attention computations does not degrade generation quality \citep{SnapKV}. Furthermore, research on decoding speedup has shown that these critical KV cache subsets are query-dependent, a phenomenon referred to as query criticality \citep{quest}. This means that each query token \( q \in Q \) corresponds to a distinct critical subset of the KV cache.
By identifying the relevant subset for the last query token in each decoding step, the decoding process can be significantly accelerated.

 However, applying query criticality to the prefilling phase introduces unique challenges.
 First, the prefilling phase already incurs significant computational overhead, and an additional step to identify critical KV cache subsets would only increase this burden.
 Second, while token-wise query criticality reduces computation via sparsification, the resulting sparsity is unstructured, as different query tokens depend on varying critical KV cache subsets. 
 This unstructured sparsity undermines the benefits of highly parallelized matrix multiplications in the prefilling phase, potentially decreasing the efficiency rather than achieving the intended acceleration.

\subsection{Locality Pattern of Query Criticality}
\begin{figure}
	\centering
	\includegraphics[height=4cm]{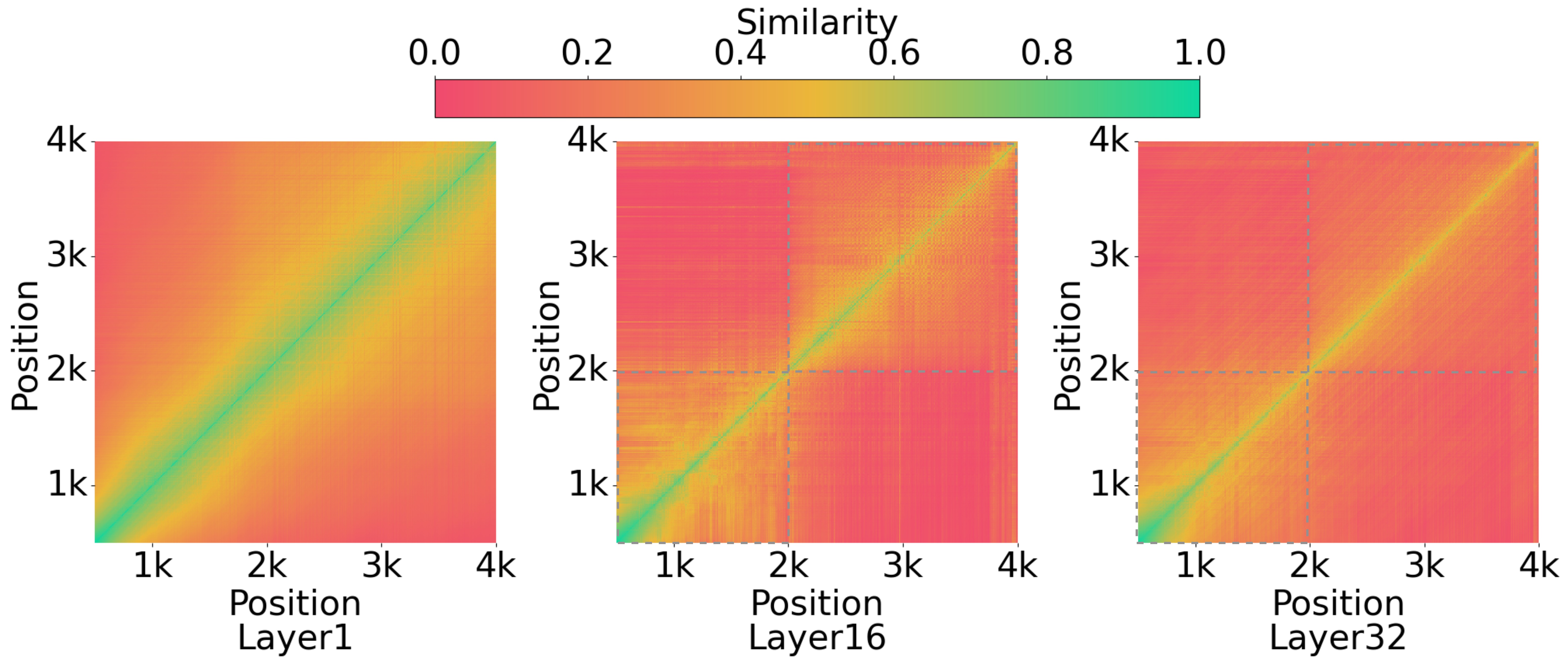}
	\caption{Locality Pattern}
	\label{fig:locality}
\end{figure}

To leverage query criticality for computation pruning, we conduct a deliberated study focusing on the long-context question answering (QA) task. Our findings reveal a locality pattern in query criticality during the prefill process. Specifically, for a 4K-length QA context on the Llama3-8B model, we first identify the critical KV cache subset \{$\hat{K}_i, \hat{V}_i$\} for each query $q_i \in Q$:
\begin{align}
\nonumber & \langle \hat{K}_i,\hat{V}_i \rangle = \langle K[index_i],V[index_i] \rangle \\
\text{where}& \: index_i  = \text{argtop}_k(\text{softmax}(q_i K^T),k=512)
\end{align}
Figure \ref{fig:locality} visualizes the similarity of critical cache subsets for different queries $q_i$ and $q_j$, using the metric $\frac{| \hat{K}_i  \cap  \hat{K}_j |}{k=512}$.
The results show that nearby queries tend to share critical KV cache subsets, as evident by the higher similarity near the diagonal. 
As the distance between queries grows, this similarity declines. 
This criticality locality pattern suggests that adjacent queries show closer criticality, allowing for collective prediction of query segments.
Thus, by leveraging the locality, it can significantly reduce the computational cost of criticality estimation, while exploiting the structured sparsity from segment-wise criticality.

\subsection{Proposed Methods}
In this section, we present CritiPrefill, a segment-wise criticality-based method for accelerating the prefilling process by leveraging the locality pattern in query criticality. 
This approach improves efficiency by estimating criticality at the segment level and utilizing structured sparsity to prune computations within the self-attention mechanism.

\subsubsection{Estimating segment-wise query ciriticality}
As described in Algorithm \ref{alg:estimation}, CritiPrefill estimates segment-wise query criticality segment-wise, significantly outperforming token-wise approaches in terms of efficiency.
First, the Query and KV Cache are firstly divided into segments and blocks. Then, representative queries and keys are selected for each segment and block via element-wise max and min operations. Lines 9-13 calculate attention scores for these representatives, producing criticality scores between segments and blocks. Additionally, a layer-fusion mechanism is employed in line 14 to refine the estimation, drawing on insights from prior works on inter-layer similarity \citep{layer_skip}.

\begin{algorithm}
\caption{Estimation of Query Criticality in One Layer}
\label{alg:estimation}
\begin{algorithmic}[1]
\STATE \textbf{Input:}  $Q, K ,V \in \mathbb{R}^{n \times d}$, query ciriticality score from last layer  $S' \in \mathbb{R}^{n_1 \times  n_2}$
\STATE \textbf{Output: } query  ciriticality score in this layer  $S \in \mathbb{R}^{n_1 \times  n_2}$
\STATE  segment number $n_1 = n / segment \: size$
\STATE  block number $n_2= n / block\: size$
\STATE $Q$ = $Q$.reshape($n_1$, $segment\: size$, $d$)
\STATE $K$ = $K$.reshape($n_2$, $block \:size$, $d$)
\STATE $Q_{max}$ = $Q$.max(dim=1); $Q_{min}$ = $Q$.min(dim=1)
\STATE $K_{max}$ = $K$.max(dim=1); $K_{min}$ = $K$.min(dim=1)
\STATE $S_1$ = softmax($Q_{max}K_{max}^T$); $S_2 $= softmax($Q_{max}K_{min}^T$)
\STATE $S_3$ = softmax($Q_{min}K_{max}^T$); $S_4 $= softmax($Q_{min}K_{min}^T$)
\STATE $S_{max} = (S_1 + S_3)/2$; $S_{min} = (S_2 + S_4)/2$
\STATE $S = \text{max}(S_{max}, S_{min})$
\STATE apply causal mask to $S$
\STATE $S$ = $\alpha S$ + (1-$\alpha$)$S'$
\STATE \textbf{Return} $S$
\end{algorithmic}
\end{algorithm}

\subsubsection{Pruning Self-attention Mechnisum}

Based on the estimated segment-wise criticality scores, CritiPrefill applies structured pruning to the self-attention mechanism (Algorithm ~\ref{alg:prun}). Rather than standard dense attention, it eliminates non-critical KV Cache blocks for the computation of each query segment, thereby reducing redundant computation operations and speeding up the prefilling phase. 
Thus, the selective attention mechanism accelerates the overall process, while maintaining generation quality.

\begin{algorithm}
\caption{Pruned Attention}
\label{alg:prun}
\begin{algorithmic}[1]
\STATE \textbf{Input:}  $Q, K ,V \in \mathbb{R}^{n \times d}$, Query Ciriticality Score  $S \in \mathbb{R}^{n_1 \times  n_2}$, Critical Budget $B$
\STATE \textbf{Output: } Attention Output $O$
\STATE  segment number $n_1 = n / segment \: size$
\STATE  block number $n_2= n / block\: size$
\STATE $Q$ = $Q$.reshape($n_1$, $segment\: size$, $d$)
    \FOR{$j = 1$ to $n_1$}
	\STATE $Q_j = Q[j]$
        \STATE critical block index $\mathcal{I} = \text{argtop}_k( S[j], B/block \:size)$
        \STATE extract critical KV cache {$\hat{K}, \hat{V}$} in cache block $\mathcal{I}$
        \STATE $A_j$ = softmax($\frac{Q_j\hat{K}^T}{\sqrt{d}})$
        \STATE  $O_j = A_j \hat{V} $
    \ENDFOR
\STATE concate all $O_j$ to $O$
\STATE \textbf{Return} $O$
\end{algorithmic}
\end{algorithm}

\section{Experiment}

In this section, we conduct experiments on datasets with various sequence lengths and scenarios to demonstrate the accuracy and acceleration effects of our approach.

\subsection{Settings}

\subsubsection{Dataset}
We conduct thorough evaluations across four datasets in both Single-hop~\citep{10.1162/tacl_a_00023} and Multi-hop QA~\citep{multi_hop1} scenarios. For Single-hop QA, where answers come from a single piece of evidence, we assess the Loogle Short-dependency QA (SD)~\citep{single_hop1} and MultiFieldQA (MF)~\citep{longbench} datasets. 
For Multi-hop QA, which requires integrating information from multiple sources, we assess the Multiple Information Retrieval (MIR)~\citep{lveval} and Comprehension and Reasoning (CR) tasks in Loogle datasets. 
Following the methodology of \citep{lveval}, we categorize each dataset into two groups based on context length: 64k and 128k for evaluation under different context lengths. Additionally, the widely-used "Needle-in-a-Haystack"~\citep{lost_in_the_middle} test, a synthetic QA task designed to assess long-context retrieval capabilities, is employed to facilitate a detailed evaluation across various context lengths.

\begin{table*}[bp]
\centering
\caption{Quality Scores on Long-Context QA Datasets}
\label{tab:quality}
\setlength{\tabcolsep}{2.4pt}
\label{tab:my-table}
\resizebox{\columnwidth}{!}{%
\begin{tabular}{lccccccccc}
\toprule
			                         & \multicolumn{4}{c}{\small Single-Hop. QA}                 & \multicolumn{4}{c}{\small Multi-Hop. QA}    \\ \cmidrule(lr){2-5}\cmidrule(lr){6-9}
                                     & \rotatebox[origin=c]{0}{\makecell{SD \\ 64k}}          & \rotatebox[origin=c]{0}{\makecell{SD \\ 128k}}    & \rotatebox[origin=c]{0}{\makecell{MF \\ 64k}} & \rotatebox[origin=c]{0}{\makecell{MF \\ 128k}}      & \rotatebox[origin=c]{0}{\makecell{CR \\ 64k}}          & \rotatebox[origin=c]{0}{\makecell{CR \\ 128k}}         & \rotatebox[origin=c]{0}{\makecell{MIR \\ 64k}} & \rotatebox[origin=c]{0}{\makecell{MIR \\ 128k}} & Ave. \\ \hline
Llama3-8B              & 41.77(1.00x)           & 40.9(1.00x)            & 26.24(1.00x)           & 24.08(1.00x)           & 17.79(1.00x)           & 18.39(1.00x)           & 18.01(1.00x)           & 18.35(1.00x)           & 25.69(1.00x)           \\
    Llama3-8B(ours w/o fusion)         & \textbf{39.72(2.22x)} & \textbf{39.41(3.38x)} & 25.35(2.49x)          & 24.05(3.94x)          & 17.67(2.43x)          & 15.87(3.66x)          & \textbf{17.31(2.44x)} & 13.94(3.68x)          & 24.16(3.03x)          \\
    Llama3-8B(ours) & 39.55(2.22x)          & 38.62(3.38x)          & \textbf{27.09(2.48x)} & \textbf{28.56(3.94x)} & \textbf{19.35(2.42x)} & \textbf{16.72(3.67x)} & 16.96(2.44x)          & \textbf{15.07(3.68x)} & \textbf{25.24(3.03x)} \\ \hline
Yi-9B                  & 31.03(1.00x)           & 16.04(1.00x)           & 14.23(1.00x)           & 12.23(1.00x)           & 10.6(1.00x)            & 6.77(1.00x)            & 8.99(1.00x)            & 5.19(1.00x)            & 13.13(1.00x)           \\
    Yi-9B(ours w/o fusion)             & 26.37(2.51x)          & 18.34(4.01x)          & 9.02(2.86x)           & 7.51(4.77x)           & \textbf{12.32(2.78x)} & 12.03(4.39x)          & 8.49(2.80x)           & \textbf{6.96(4.42x)}  & 12.63(3.56x)          \\
    Yi-9B(ours)     & \textbf{27.89(2.52x)} & \textbf{22.19(3.98x)} & \textbf{9.59(2.86x)}  & \textbf{8.32(4.71x)}  & 11.16(2.79x)          & \textbf{12.34(4.35x)} & \textbf{9.11(2.80x)}  & 4.56(4.37x)           & \textbf{13.14(3.55x)} \\
\toprule
\end{tabular}%
}
\end{table*}

\subsubsection{Baseline}

We compare the CritiPrefill method with vanilla LLMs to demonstrate that our approach significantly accelerates prefilling with minimal quality loss. Additionally, a CritiPreill method without the layer-fusion mechanism is also included as an ablation study to illustrate that this mechanism effectively enhances the accuracy of segment-wise criticality estimation. All methods are based on two open-source LLMs, Llama3-8B-1M \citep{llama2}, Yi-9B-200K~\citep{yi}, which are widely used for long context evaluation due to their moderate parameter size and remarkable capability of long context processing\citep{xiong2023effective}.

\subsubsection{Parameter Settings}

In all experiments, the query segment size is set to 512, the key-value block size to 32, and the budget size to 1024. The layer-fusion hyperparameter, $\alpha$, is fixed at 0.25. All experiments employ Flash Attention \citep{fa2} and FP16 for efficient computation on Nvidia A100 GPUs. For more details, please refer to our code at https://github.com/66RING/CritiPrefill.

\subsection{Evaluations cross multiple scenarios}

The test results for each model across all datasets are summarized in Table \ref{tab:quality}. Overall, our CritiPrefill method achieves an average speedup of over 3x across all datasets, with minimal degradation in quality. Specifically, on the Llama3-8B model, CritiPrefill achieves a 3.0x speedup with only a slight drop in quality score from 25.69 to 25.24. The Yi-9B model attains a 3.4x speedup, while the quality score remains nearly unchanged, shifting from 13.13 to 13.14. This demonstrates that CritiPrefill's segment-wise estimation algorithm not only effectively approximates query criticality but also successfully implements structured attention pruning, resulting in tangible acceleration gains. Furthermore, when examining datasets with varying sequence lengths, CritiPrefill's acceleration becomes more pronounced with longer sequences. For instance, on the SD dataset using Llama3-8B, CritiPrefill achieves speedups of 2.2x and 3.3x at the 64K and 128K sequence lengths, respectively. This is primarily because, with longer sequences, CritiPrefill can prune more irrelevant operations during AttentionPrefill, leading to greater performance improvements.

Furthermore, comparing CritiPrefill with CritiPrefill without layer-fusion shows that while there is almost no difference in speed, our method without layer-fusion shows a significantly lower generation quality. This indicates that the layer-fusion mechanism effectively enhances the accuracy of criticality estimation without introducing excessive computational overhead.

\subsection{Evaluations on Needle-in-a-Haystack Test}

\begin{figure}[t]
    \centering
    \begin{subfigure}{0.8\linewidth} 
        \centering
        \includegraphics[width=\linewidth]{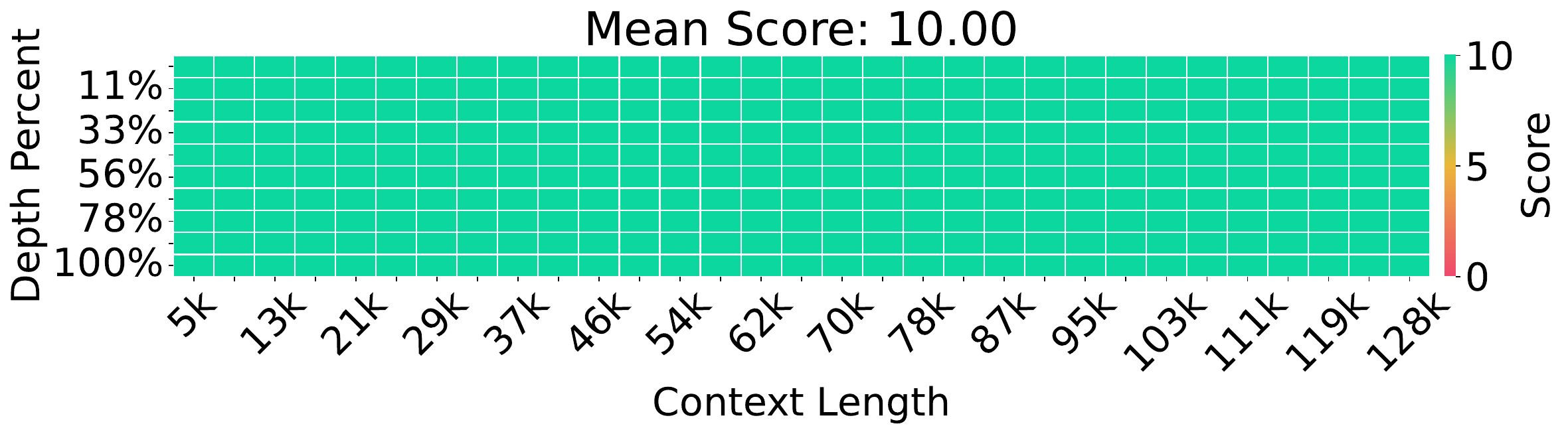}
         \vspace{-15px}
        \caption{Llama3-8B}
        \label{fig:llama3_needle_ours}
    \end{subfigure}
    \hfill
    \begin{subfigure}{0.8\linewidth} 
        \centering
        \includegraphics[width=\linewidth]{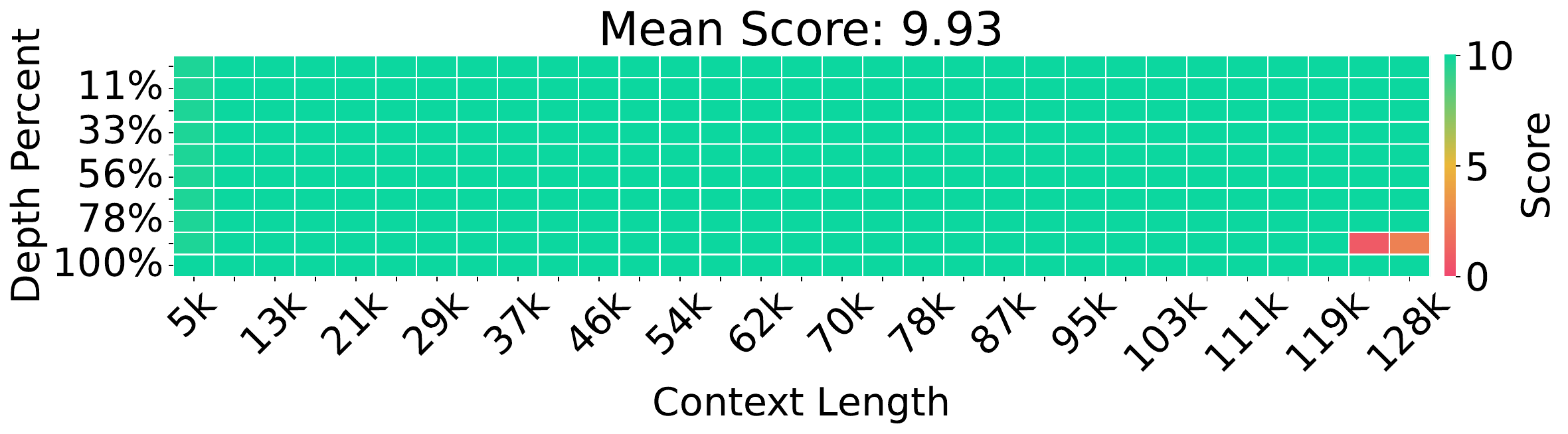}
        \vspace{-15px}
        \caption{Yi-9B}
        \label{fig:yi_needle_ours}
    \end{subfigure}
    \caption{CritiPrefill on Needle-in-a-Haystack Test. 		
		This test involves inserting an answer within a large context and evaluating the retrieval ability in response to the corresponding question. The Average Score is derived from averaging scores across various context lengths (x-axis) and insertion depths (y-axis), with a maximum score of 10.    
}
    \label{fig:needle}
    \vspace{-10px}
\end{figure}

\begin{figure}
	\centering
	\includegraphics[height=4.5cm]{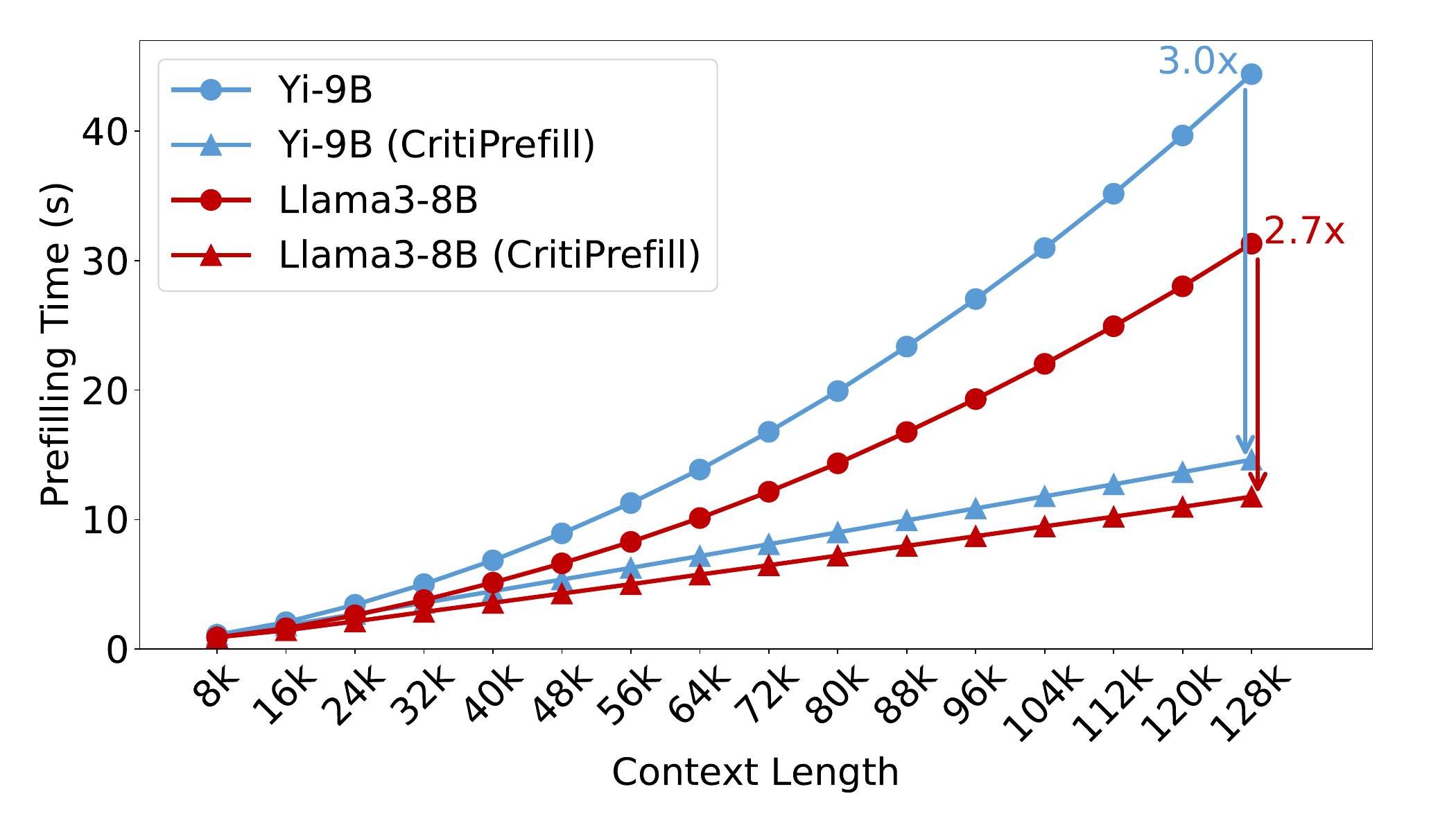}
	\caption{Prefilling Time on Need-in-a-Haystack Test}
    \label{fig:speedup_ratio}
    \vspace{-10px}
\end{figure}

The Needle-in-a-Haystack test, as shown in Figure~\ref{fig:needle}, is conducted to thoroughly evaluate the impact of the information retrieval capabilities across varying lengths. Both the Llama-3 and Yi models exhibit nearly lossless performance in the Needle-in-a-Haystack test, ranging from 0 to 128K context lengths when utilizing CritiPrefill methods. As illustrated in Figure~\ref{fig:speedup_ratio}, the acceleration provided by CritiPrefill becomes increasingly significant as sequence lengths grow. It effectively reduces the quadratic prefilling time to linear without compromising performance at lower sequence lengths like 8K-16K.

\section{Conclusion}
In this work, we propose CritiPrefill, a plug-and-play method to accelerate the prefilling phase in LLMs. Our key insight is identifying a locality pattern where neighboring query tokens rely on similar critical KV cache. This enables efficient segment-wise criticality estimation and structured pruning of self-attention computation, significantly reducing the computational load during the prefilling phase. By introducing a layer-fusion mechanism to refine criticality across layers, CritiPrefill enhances generation quality without compromising efficiency. Experiments on long-context QA tasks demonstrate up to 3.0x speedup with minimal accuracy loss on models such as Llama3-8B and Yi-9B. CritiPrefill provides an effective solution to the long-sequence inference bottleneck, requiring no architectural changes or fine-tuning, making it highly adaptable for practical applications.

\bibliography{iclr2025_conference}
\bibliographystyle{iclr2025_conference}

\end{document}